# Comparing Optical Flow and Deep Learning to Enable Computationally Efficient Traffic Event Detection with Space-Filling Curves

Tayssir Bouraffa[1], Elias Kjellberg Carlson[1], Erik Wessman[1], Ali Nouri[1,2], Pierre Lamart[3], Christian Berger[3]

*Abstract*— Gathering data and identifying events in various traffic situations remain an essential challenge for the systematic evaluation of a perception system's performance. Analyzing large-scale, typically unstructured, multi- modal, time-series data obtained from video, radar, and LiDAR is computationally demanding, particularly when meta-information or annotations are missing. We compare Optical Flow (OF) and Deep Learning (DL) to feed computationally efficient event detection via space-filling curves on video data from a forward-facing, in-vehicle camera. Our first approach leverages unexpected disturbances in the OF field from vehicle surroundings; the second approach is a DL model trained on human visual attention to predict a driver's gaze to spot potential event locations. We feed these results to a space-filling curve to reduce dimensionality and achieve computationally efficient event retrieval. We systematically evaluate our concept by obtaining characteristic patterns for both approaches from a large-scale virtual dataset (SMIRK) and applied our findings to the Zenseact Open Dataset (ZOD), a large multi-modal, real-world dataset, collected over two years in 14 different European countries. Our results yield that the OF approach excels in specificity and reduces false positives, while the DL approach demonstrates superior sensitivity. Both approaches offer comparable processing speed, making them suitable for real-time applications.

## I. INTRODUCTION

The identification of anomalous behaviour in traffic scenarios is a crucial task for safe driving, particularly in vehicles equipped with Advanced Driver Assistance Systems (ADAS). Potential dangers can appear during crowded traffic, such as a sudden pedestrian or motorcycle crossing or unexpected vehicle overtaking, which necessitates a high vigilance of the driver. Hence, implementing a robust system that can detect such events in front of the vehicle can assist and warn the driver when a potential danger occurs.

As vehicles become more autonomous, the reliance on accurate and reliable detection systems becomes critical, particularly in unsupervised Autonomous Driving Systems (ADS), where perception relies entirely on the system without human supervision. In such systems, any false positives or negatives in the detection of objects like vehicles and pedestrians can lead to safety-critical incidents. Thus, the reliability and safety of the perception system become crucial and to address this, multiple sensor technologies are employed. Each technology independently perceives the environment and their data are then fused to more accurately model it. By leveraging the diverse strengths and mitigating the weaknesses of each sensor technology, the implementation enhances the overall reliability and safety of the autonomous driving system.

### A. Problem Statement and Motivation

In recent years, ADAS have impressively advanced through the incorporation of multi-model sensors, including rear and side radar, dashboard cameras, and backup cameras, while LiDAR is increasingly added to enhance ADS. The information collected through these sensors is used to detect vehicles and pedestrians, where the ego-vehicle becomes aware of its surroundings.

Autonomous vehicles rely solely on sensor inputs, without human intervention, necessitating a diverse array of sensors to ensure robust perception. Each sensor type, whether radar, LiDAR, or cameras, individually interprets the environment, with their perceptions subsequently cross-validated against one another to enhance reliability. This inter-sensor validation ensures system confidence is based on their collective accuracy. While radar and LiDAR are valuable for their range and resistance to environmental influences, including cameras is essential. Cameras, being cost-effective and passive, complement the active sensors by providing critical visual data, making the overall system more reliable and comprehensive. Hence, we focus in our work on event identification from a forward-facing camera in a moving car. The main challenge originates from the dynamic nature of the background, the moving camera, and the dynamic behavior of the surrounding vehicles and pedestrians.

Vehicles that are equipped with multiple sensors can easily generate hundreds of megabytes and even up to gigabytes of data per second. While such data collections are essential during the development and especially for the validation of perception stacks, naïvely processing all data is not effective and hence, computationally efficient approaches to retrieve relevant events or interesting traffic scenarios are needed. In order to support such information processing, motion detection is a computer vision technique that can be used to detect an object against its background. This can be achieved through the implementation of Optical Flow algorithms, such as Farnebäck [1], and Lucas-Kanade [2], enabling the estimation of moving regions or pixels between consecutive frames [3] and in-turn, allow the extraction of relevant features

*This work has been partially supported by the European research project "SUNRISE" funded by the European Union's Horizon Europe Research & Innovation Actions under grant agreement No. 101069573. This work has been partially supported by the Wallenberg AI Autonomous Systems and Software Program (WASP) funded by the Knut and Alice Wallenberg Foundation.

[1]Chalmers University of Technology, Department of Computer Science and Engineering, Gothenburg, Sweden `tayssir@chalmers.se`
[2]Volvo Cars, Gothenburg, Sweden `ali.nouri@volvocars.com`
[3]University of Gothenburg, Department of Computer Science and Engineering, Gothenburg, Sweden `{pierre.lamart,christian.berger}@gu.se`

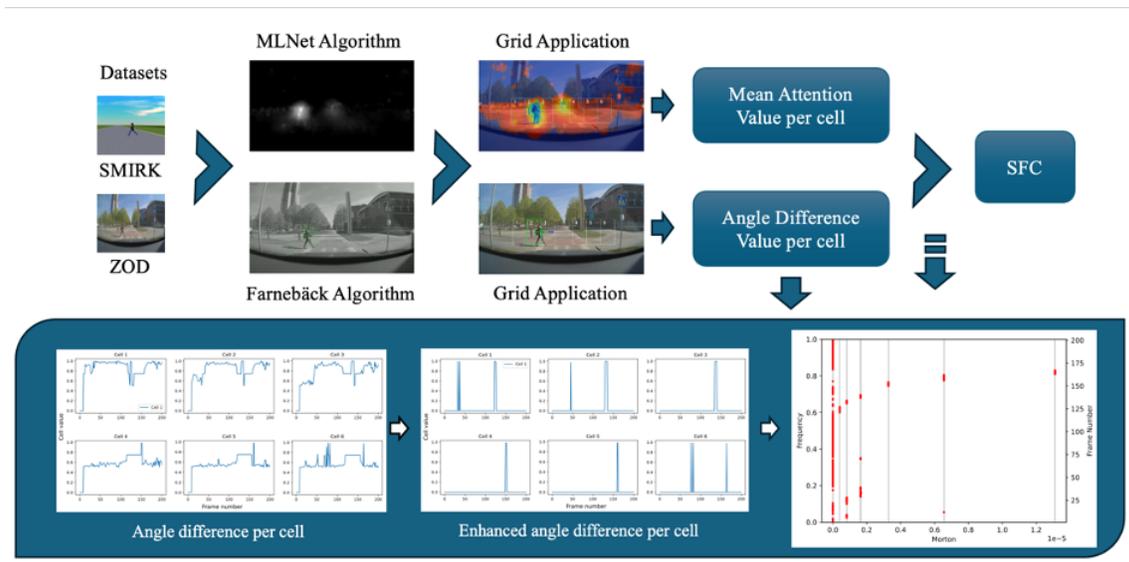

Fig. 1. The overall workflow depicting the concept of event retrieval from unstructured video data: Beginning with the SMIRK and ZOD datasets, the MLNet and Farnebäck algorithms extract deep learning and optical flow features. A grid is then applied to these features to calculate the mean attention values or mean flow vectors for each cell, respectively. Leveraging SFCs for dimensionality reduction results in stripe plots. Events are subsequently identified by analyzing spatiotemporal relationships within these stripe plots, thus enabling the interpretation of complex, multi-dimensional time-series data.
* At the bottom box, only the OF-SFC variant is depicted for the evaluation part.

to support event identification. Alternatively to such analytical approaches, deep learning models such as Convolutional Neural Network (CNN), Deep Multi-Level Network (MLNet), and Reinforcement Learning (RL) techniques can be leveraged to effectively model a driver's visual attention that in-turn allows extracting of relevant features to support event identification.

Uneven road surfaces, abrupt speed changes, or aggressive maneuvering play a critical role in the reliability of autonomous driving technologies. Such factors influence vehicle dynamics significantly, like pitching from acceleration or braking, which can disrupt the accuracy of OF and visual attention. Understanding and accounting for these disturbances is crucial for accurate motion pattern analysis and attention prediction within a driving environment.

### B. Research Goal and Research Questions

The research goal of this paper is to systematically compare the applicability of both, OF-based and DL-based feature extractors to obtain semantically relevant, multi-dimensional, time-series data from video sequences that in-turn will be transformed into single-dimensional time-series via space-filling curves (SFCs) to allow computationally efficient event identification. We derive the following research questions:

**RQ-1:** What does a conceptual framework and prototype to systematically transform unstructured, time-series data into semantically relevant, multi-dimensional, time-series data suitable to be fed into SFCs for computationally efficient event identification look like?

**RQ-2:** What is the qualitative and quantitative performance of the aforementioned framework?

### C. Contributions

Our main contributions include the design of a conceptual framework to extract semantically relevant features from unstructured, time-series data suitable to be fed into SFCs to allow for computationally efficient event identification. Firstly, we identify semantically relevant features to be extracted for one OF algorithm and one DL model by systematic experimentation on a synthetic dataset in a simulation. Afterwards, we transfer the identified parameters from the controlled environment to a real-world dataset to confirm our observations and to evaluate the qualitative (sensitivity, specificity, mean IoU, and F1 scores) and quantitative (processing times and scalability) performance of the proposed concept. We adopted Design Science Research for our methodology based on recommendations of Hevner et al. [4] to systematically conduct the evaluation.

### D. Scope

We focus in our work on pedestrian and cyclist crossings captured by a monocular, forward-facing color camera in urban environments. We parameterize our approach by using a synthetic dataset and assess the approach afterwards with an automotive dataset created from industry. We selected 84 scenarios with pedestrian and cyclist crossings from left to right or vice versa for the evaluation with the real-world data.

### E. Structure of the Article

The remainder of this paper is organized as follows: Section II introduces the OF, visual attention, and SFCs, providing a brief review of relevant research in event detection. Section III outlines the proposed approach, emphasizing its primary contributions. Section IV describes the methodology for evaluation, detailing the experimental setup. Section

V discusses and analyzes the experimental results. Finally, the conclusion is summarized in Section VI.

## II. RELATED WORK

In previous work to extract the moving vehicles from video sequences, different methods have been employed, including vehicle tracking via bounding boxes, vehicle recognition using deep learning techniques, and vehicle detection based on their appearance [5]. While these algorithms perform well with stationary cameras, they encounter significant challenges when applied to cameras mounted on moving vehicles. These challenges are mainly due to the dynamic nature of the background and the speed variation of the surrounding vehicles.

An effective approach that takes advantage of the scene dynamics is the deployment of motion detection-based algorithms. Different methods have been explored to detect motion in a driving video, such as frame difference, temporal difference, background subtraction, and OF algorithms. Optical flow estimation is widely used in motion detection to obtain velocity information. The optical flow is determined on the assumption of brightness constancy, which posits that the timestamps of two successive frames are in close proximity and the brightness of the same locations in the real world is constant [6].

Alonso et al. demonstrated that by calculating OF vectors on each frame, a decision support system can determine potential lane-changing maneuvers by the tracked vehicles [7]. Yuan et al. proposed an anomaly detection approach in traffic scenes that takes into account both motion orientation and magnitude of the flow vectors, the resulting feature maps are fused via a Bayesian model for more robust results [8]. Additionally, Kilicarslan et al. suggested an approach that can detect dangerous events by analyzing the divergence of motion both vertically and horizontally within driving videos [9]. Ramirez et al. proposed an approach that integrates both appearance and motion cues, where forward and backward OFs are calculated to effectively detect overtaking vehicles [10]. Building on these foundations, a motion detection approach is adopted in this paper to identify semantically relevant regions of interest by selecting motion patterns that deviate from expected ones.

In the past decade, saliency prediction has received significant attention. Several studies have identified the initial fixation point, a human primary focus upon first viewing an image. Other researchers have concentrated on identifying and emphasizing the regions containing the most significant objects within an image. Cornia et al. introduced a deep learning architecture designed to identify saliency regions, leveraging multi-level feature extraction capabilities of CNNs [11]. Concurrently, the study of drivers' visual attention in traffic scenarios has advanced. Deng et al. developed a convolutional-deconvolutional neural network model that predicts a driver's eye fixations, incorporating both bottom-up and top-down information pertinent to traffic driving, and trained with eye-tracking data from multiple drivers [12]. Similarly, Bao et al. introduced a RL approach to emulate human visual attention mechanisms for anticipating traffic accidents [13]. Additionally, Xia et al. proposed a driver attention prediction model tailored for critical scenarios, adept at recognizing sophisticated behaviors such as the attention given to crossing pedestrians and cyclists [14].

In dealing with large-scale automotive scenarios, computationally efficient approaches are needed to handle such multi-dimensional data. Perez et al. proposed a brute-force method for extracting automotive events from naturalistic driving data [15]. Bader pointed out that SFCs are suitable for dimensionality reduction, a method that performs data linearization to transform multi-dimensional data into a single-dimensional representation [16]. Hulbert et al. explore the potential of applying SFC to scalable datasets [17]. Our previous works (cf. [18], [19] demonstrate the effectiveness of using SFCs in retrieving events in the automotive context for structured, time-series data. However, exploring the feasibility of this method and applying it to video data remains an open research area that we address in this work.

## III. PROPOSED METHOD

We propose a concept to process unstructured video sequences to extract semantically relevant, multi-dimensional, time-series data that in-turn will be transformed into single-dimensional time-series via SFCs to allow for computationally efficient event identification and retrieval as shown in Fig. 1. The computational efficiency in the final step relies on a theoretical property of SFCs to preserve locality of tuples in the multi-dimensional space after transformation into the single-dimensional space (cf. Bader, [16]). As we have shown in Berger and Birkemeyer (cf. [18]), the elements in the single-dimensional space correlate in their spatial distribution, spread, and temporal occurrence semantically with events in the original, multi-dimensional space. We exploited this correlation for efficiently searching for events of interest to find harsh braking maneuvers or lane-changes (cf. [18]) or passing maneuvers through roundabouts (cf. [19]) by solely analyzing SFC-transformed accelerations over time.

We are expanding the aforementioned approach that focuses on vehicle kinematics by integrating information from a vehicle's surrounding like from cameras. However, to preserve the benefits of SFCs, such unstructured data needs to be transformed while retaining semantically essential information as SFCs expect structured and multi-dimensional input vectors. We use and compare two conceptually different approaches to extract semantically relevant features from unstructured video data in preparation for feeding them into SFCs: The first variant, which we denote as OF-SFC, uses an OF algorithm to extract image sections in comparison to the background, and the second variant that we label CNN-SFC employs a CNN to extract multi-level features to predict saliency regions.

Both variants are semantically filtered by a domain-specific region-of-interest (RoI) grid to obtain the input-vectors for the SFC. Fig. 2 illustrates the semantic feature extraction stage using one scenario from the test dataset. We define this RoI within a video frame to declare specific areas where relevant events are most likely to occur based on domain knowledge. We intentionally exclude regions prone to distortions from surrounding elements such as bridges, buildings, and trees, as well as vehicle dynamics-imposed noise caused by pitching or rolling. It is noteworthy that the entire field of view is affected by the vehicle dynamics to some degree. When pitching occurs, for example during acceleration or braking, the vehicle's forward or backward motion induces vertical displacement in the camera's field of view. This results in objects within the scene appearing to move upward or downward. The selected RoI reduces the influence of these effects in the feature extraction stage, enabling a stronger semantic connection among the different cells. Complementary thereto, a human driver usually focuses towards a traffic scene's vanishing point, creating a constant area of attention at this location for the CNN predicting the saliency map. However, this area rarely contains any significant domain-specific information when considered in isolation. Therefore, we exclude it to reduce such unwanted noise for the saliency attention. As a result, our RoI is located centrally within a frame denoting our domain-specific area for events relevant for traffic situation such as pedestrian crossings.

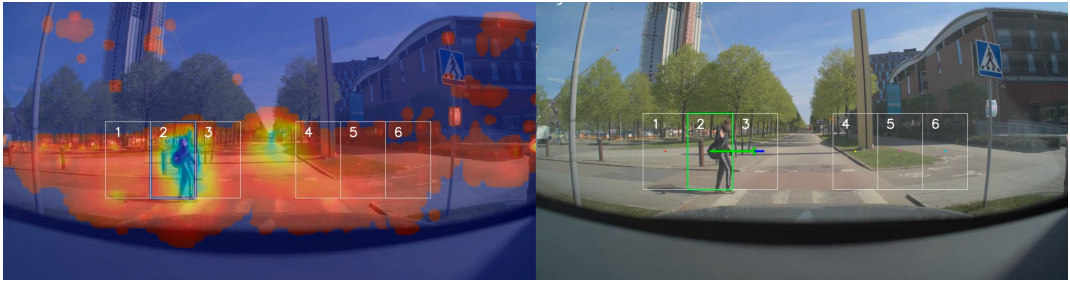

Fig. 2. The variants CNN-SFC (on the left) and OF-SFC (on the right) applied to a traffic event where a pedestrian is crossing. The CNN-SFC is computing a saliency map predicting human attention, while the OF-SFC is computing disturbances in the vector field compared to the generally expected optical flow.

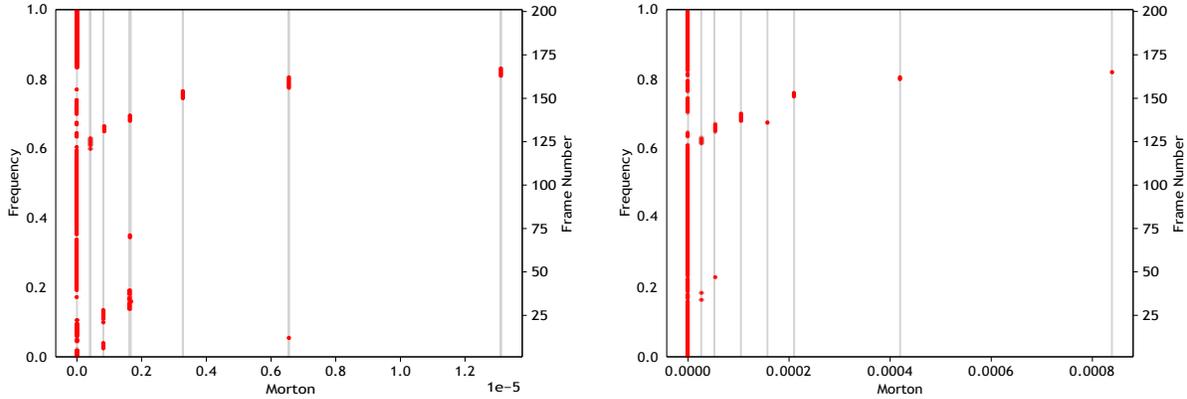

Fig. 3. Corresponding stripes plot after feeding the multi-dimensional vectors obtained from the variants CNN-SFC (on the left) and OF-SFC (on the right) to the Z-order space-filling curve. The overlay dots represent the temporal correlations between the pedestrian crossing a street and the respective variants follow that event in their overlay grids to obtain the multi-dimensional feature vector: The overlay points show a trend in the spatiotemporal domain of the SFC that we exploit for traffic event identification.

### A. Semantic Feature Extraction using Optical Flow

The conceptual idea behind the OF-SFC variant is to detect areas within a video frame where the motion of pixel regions deviates from the typically expected pattern over time and hence, highlighting potential areas of interest in a traffic scene. We chose Farnebäck's algorithm to calculate the OF because it has been shown to be more suitable for complex and dynamic backgrounds typical of real-world driving footage [8]. In contrast thereto, the CNN-SFC variant aims to infer the visual attention within a frame by highlighting saliency regions over time. The OF-SFC variant does not require any human supplied annotation and training compared to machine learning (ML), while the CNN to obtain saliency maps require annotations and domain-specific training.

The OF-SFC variant examines vector flow fields within the RoI. We extract potential traffic events by comparing the vector flow fields against the typically expected flow induced by the vehicle's own movement resulting in flow disturbances. In scenarios such as an overtaking vehicle or a pedestrian crossing, events can be characterized by clusters of pixels whose vector flow field is moving in a different direction than they are expected compared to the focus-of-expansion (FoE) or the vanishing point. As an example, the illustration to the right in Fig. 2 is spotting the vector field movement caused by a crossing pedestrian (green vector).

We discretize the vector field flow using a domain-specific grid laid over the RoI to segment the image into semantically meaningful grid cells (denoted as 1 to 6 in Fig. 2. We compute the mean flow vector within each grid cell representing the average motion within that cell *i*. Specifically, for a given mean flow vector at a pixel point $(x, y)$ in a frame $t$, $v(x, y)$ and $u(x, y)$ represent the vertical and horizontal components of the mean flow vector, respectively. The direction of apparent motion at this pixel can be deduced from the angle of the mean flow vector. At the current frame $t$, this angle should approximate an event angle $\alpha$ for a pedestrian crossing from left to right and $-\alpha$ for a pedestrian crossing from right to left. The acceptable range for these event angles is defined by the threshold $\vartheta$. By comparing the direction of the average mean flow vector from the previous $t-n$ frames with the direction of the average mean flow vector from the previous $t-m$ frames, any unexpected changes in the flow motion relative to its expected pattern can be identified. The group of pixels within a frame is identified as part of an event when the angle difference between the previous $n$ average mean flow vector and the $m$ average mean flow vector of the preceding frames exceeds a predefined threshold $\delta$. In the end, we obtain a multi-dimensional vector composed of the angle differences between mean flows from each cell over time.

*B. Semantic Feature Extraction using CNN-based Saliency Maps*

Inspired by the perception mechanism of the human visual system, a deep learning architecture is modeled for predicting saliency maps. The saliency model is adopted to learn the visual attention behavior to address where traffic situations such as pedestrian crossing or cut-in events may occur. We apply the MLNet algorithm [11] in the CNN-SFC variant across an entire video frame to comprehensively observe the scene, leveraging the attention mechanism to identify pertinent driving events. The process begins with an input image, from which a CNN based on the VGG-16 architecture [20] computes features at low, medium, and high levels. The extracted feature maps are subsequently fed into an encoding network tasked with learning a feature weighting function, which is crucial for generating feature maps that are specifically tailored for saliency detection. Following this, a prior image is learned and strategically applied to enhance the accuracy of the resulting saliency map. Saliency maps are generated for each video frame where a pixel's intensity is reflecting the level of saliency.

We also discretize the RoI to facilitate the analysis of visual attention over time, calculating the mean saliency value within each cell grid. This allows us to examine the distribution and intensity of visual attention across a frame. By tracking how these mean saliency values evolve across the cell grid, we can identify regions of specific visual interest. A saliency region is identified as part of an event when the mean saliency of that cell exceeds a predefined threshold $\gamma$, and only the cell with the highest mean saliency value within this threshold is designated as active. As an example, the illustration to the left in Fig. 2 is highlighting a crossing pedestrian draw attention from the CNN and resulting in an intense saliency. In result, we obtain a multi-dimensional vector composed of the mean attention value from each cell.

*C. Dimensionality Reduction with Space-Filling Curves*

The next step in our proposed concept as shown in Fig. 1 is to feed the multi-dimensional feature vectors over time to a SFC. The motivation for this step is to enable computationally efficient event identification and retrieval at scale as we have already demonstrated in our previous work focusing purely on vehicle kinematics (cf. [18]). For the OF-SFC variant, individual vectors in the multi-dimensional space denote the angle differences within each grid cell of the RoI, and for the CNN-SFC variant, the individual vectors represent the mean attention value within each grid cell in the RoI, respectively. We decided to use the Z-order curve to transform the multi-dimensional vectors into their corresponding single-dimensional representations as our previous experiments have shown this curve's domain-specific suitability (cf. [19]). The resulting single-dimensional representations, the so called Morton codes, generate characteristic stripe patterns (CSPs) as illustrated in Fig. 3. These CSPs correlate semantically in their distribution, spread, and temporal occurrence with events in the multi-dimensional space, ie., in our case with traffic events of interest such as a crossing pedestrian as shown in Fig. 2. We exploit these spatiotemporal patterns, ie., the trend with the red dots laid over to the stripes in Fig. 3, to enable event identification. Different traffic situations are characterized by distinct spatiotemporal patterns in their CSP. For instance, the pattern semantically linked with a pedestrian crossing event differs from a vehicle cut-in scenario on a highway.

> *Answering RQ-1:* We have proposed two variants in our concept to transform data from unstructured video sequences into structured, multi-dimensional feature vectors: The first is based on non-annotated, optical flow, and the second is using a CNN to predict human attention. We feed these feature vectors into the Z-order space-filling curve to obtain characteristic stripes resulting in spatiotemporal patterns correlating with specific traffic situations.

IV. RESEARCH METHODOLOGY FOR EVALUATION

We adopted Design Science Research as proposed by Hevner et al. [4] to evaluate the qualitative and quantitative performance of our proposed approach for computationally efficient event identification. We used two publicly accessible datasets: SMIRK [21] and the Zenseact Open Dataset (ZOD) [22]. SMIRK is a synthetic dataset that comprises 5,028 scenarios in total, from which 4,928 scenarios cover crossing scenarios for pedestrians, and 100 scenarios represent non-pedestrian crossings; the 4,928 scenarios include 616x4 male pedestrian, 616x3 female, and 616 child crossing scenarios. Each subset of 616 scenarios is further categorized by the direction of movement: 280 involve pedestrians crossing from left to right, another 280 from right to left, and 28 moving parallel to the road towards the ego vehicle, with an additional 28 moving away [21]. We decided to use SMIRK for two main reasons: Firstly, a synthetic dataset created in a virtual environment allows controllability of potentially influential factors such as a vehicle's rolling or pitching; secondly, we can also easily get annotations to semantically correlate spatiotemporal patterns from our results CSPs after transforming the multi-dimensional data via a SFC into their corresponding single-dimensional representations. The SMIRK dataset allowed us to confirm that our approach is allowing us to identify semantically relevant traffic situations.

TABLE I
Performance of OF-SFC and CNN-SFC variants on SMIRK dataset.

| Variant | F1 Score | Sensitivity | Specificity | Mean IoU | Speed (FPS) |
|---|---|---|---|---|---|
| OF-SFC | 0.808 | 0.698 | 0.996 | 0.732 | 21.88 |
| CNN-SFC | 0.737 | 0.630 | 0.988 | 0.705 | 28.26 |

TABLE II
Performance of OF-SFC and CNN-SFC variants on ZOD dataset (selection of pedestrian crossing and non-pedestrian crossing scenarios).

| ZOD sub-dataset | Variant | F1 Score | Sensitivity | Specificity | Mean IoU | Speed (FPS) |
|---|---|---|---|---|---|---|
| $ZOD_1$ | OF-SFC | 0.455 | 0.313 | 0.941 | 0.514 | 21.05 |
|  | CNN-SFC | 0.621 | 0.643 | 0.684 | 0.499 | 20.71 |
| $ZOD_2$ | OF-SFC | 0.455 | 0.313 | 0.941 | 0.514 | 21.00 |
|  | CNN-SFC | 0.643 | 0.643 | 0.737 | 0.499 | 21.24 |
| $ZOD_3$ | OF-SFC | 0.417 | 0.313 | 0.824 | 0.514 | 20.92 |
|  | CNN-SFC | 0.581 | 0.643 | 0.579 | 0.499 | 21.43 |
| $ZOD_4$ | OF-SFC | 0.435 | 0.313 | 0.882 | 0.514 | 20.97 |
|  | CNN-SFC | 0.581 | 0.643 | 0.579 | 0.499 | 21.08 |

We use the ZOD dataset to evaluate the transferability of our results obtained from the synthetic dataset to real world data, where artifacts from non-controllable vehicle kinematics such as pitching and rolling, as well as noise after applying OF-SFC and CNN-SFC are unavoidable. ZOD is a real-world dataset that encompasses 1,473 driving scenarios of 20s length each [22]. From the ZOD dataset, we filtered 358 scenarios containing any pedestrian and selected thereof 16 scenarios where a pedestrian was crossing the street clearly (either from left-to-right in 11 cases, or from right-to-left in 5 cases). We sampled another 68 scenarios where a pedestrian was not crossing completely, not at all, or using a bridge for crossing for example. We constructed 4 combinations of this subset for our qualitative evaluations, each subset includes the 16 videos that contain a pedestrian clearly crossing and another 17 videos that do not contain a relevant crossing. This allowed us to create 4 evaluation groups to qualitatively assess the robustness of the event detector on the real-world data. The subset of 84 scenarios (16 positive crossing cases and 68 no-crossing cases) show daylight time driving under sunny or cloudy weather conditions. The driving scenarios were collected in Sweden, France, and Germany. The complete list of selected scenarios is available here to support replicability of our study (cf. below) GitHub repository.

To align SMIRK and ZOD, we manually annotated the pedestrian crossings in both datasets to define an event window that we were interested to obtain with both, OF-SFC and CNN-SFC. The event began the moment a pedestrian was stepping on the road until either the video sequence came to an end or the pedestrian left the road again. We use these annotations as ground truth to qualitatively evaluate the performance of the event detection after the final stage of OF-SFC and CNN-SFC when filtering the single-dimensional data in the stripes plot according to the spatiotemporal relations that we identified by using the synthetic data from SMIRK. We evaluate the quantitative performance of our approach by comparing the computational duration of the frame processing. We conducted our experiments using Ubuntu 23.10 running on a $12^{th}$ Generation Intel(R) Core(TM) i7-12700K CPU with 32GB DDR5 RAM, a 512GB M.2 SSD supported by an NVIDIA GeForce RTX 3080 GPU.

We configured the Farnebäck algorithm to compute a dense OF with specific parameters targeting traffic events such as pedestrian crossings and vehicle cut-ins. These parameters include a pyramid scale of 0.5, 3 pyramid levels, and the averaging window size of 15. The algorithm leverages a pixel neighborhood size of 5 for polynomial expansion and a Gaussian standard deviation of 1.2 in the polynomial expansion to smooth the derivatives. The algorithm iterates 3 times at each pyramid level to refine the flow vectors. These parameters are chosen to emphasize the detection of large and fast motions, crucial for identifying pedestrian crossings while de-emphasizing minor details and slower motions.

We tailored the MLNet algorithm with specific parameters to support predicting the human attention in traffic events of interest. MLNet is built upon the VGG-16 model, utilizing three convolutional layers conv3, conv4, and conv5 while removing the last max pooling layer. Each layer uses a kernel size of 5, a stride of 1, and padding of 2, maintaining the same spatial size across layers. These layers are concatenated to form a tensor with 1280 channels, which is then input to a dropout layer with a retain probability of 0.5 to enhance generalization. Subsequently, a convolutional layer processes this tensor to learn 64 saliency-specific feature maps using a 3x3 kernel. A final 1x1 convolutional layer is applied to weight the importance of each feature map, producing the final predicted feature map. Following the DRIVE model, the saliency module is trained on the fixation data of the DADA-2000 training set [13]. We share the implementation as open source available here GitHub repository.

## V. Results and Discussion

We present the results from our evaluation in Table I for the synthetic SMIRK dataset, and in Table II for the real world ZOD dataset. To assess the efficacy of the two variants, we include evaluation metrics, such as F1 score, sensitivity, specificity, and mean intersection-over-union (IoU). Additionally, we report the total processing time per frame as frames per second (FPS) to evaluate their efficiency.

### A. Detecting Pedestrian Crossings using OF-SFC Variant

The Farnebäck algorithm was computed on a grid, segmented into 6 cells designated for pedestrian crossing as depicted in Fig. 2 (right). Notably, the pedestrian crossing can typically emerge within the scene from left to right or from right to left. For each cell, we compute the angle difference between the average mean flow vector of the previous $n = 4$ frames and the average mean flow vector of the preceding $m = 7$ frames. After setting the angle difference threshold $\delta = 75$, the event angle $\alpha = 90$, and the event angle range threshold $\vartheta = 20$, peak values indicating the presence of a pedestrian crossing are retained, while other values are reduced to zero and hence, resulting in a clearly visible mean motion vector (depicted in green in Fig. 2).

### B. Detecting Pedestrian Crossings using CNN-SFC Variant

The MLNet algorithm was applied to each video frame and similar to the OF-SFC variant, the grid was divided into 6 cells, with 3 on the left side and 3 on the right side, with a gap in the center to exclude the attention region constantly influenced by the predicted driver's focus on the FoE. In this approach the mean attention value of each cell grid is computed. Thresholds are set at $\gamma = 0.2$ for SMIRK and $\gamma = 0.35$ for ZOD to identify pedestrian crossings indicated by subsequent peak values throughout the grid cells as illustrated in Fig. 2. The resulting six mean attention values from each cell serve as an multi-dimensional input for the Z-order curve. For a pedestrian crossing to be confirmed, at least three cells must be sequentially activated with at least one cell activated on each side, and there should be no sudden jump exceeding two cells.

Table I illustrates a comparative performance of OF-SFC and CNN-SFC variants, as applied to the SMIRK dataset. The OF-SFC variant exhibits a higher F1 Score, indicating that OF-SFC has a superior balance of precision and sensitivity in this context. Sensitivity, or true positive rate, is also higher for OF-SFC, suggesting that it is more effective at correctly identifying true events than CNN-SFC. In terms of specificity, which measures the true negative rate, both methods show high performance. This high specificity indicates that both variants are adept at correctly identifying non-events, although OF-SFC has a slight edge. The mean IoU is a metric used to evaluate the accuracy of an object detector on a particular dataset, higher values indicate better performance. Here, OF-SFC has a mean IoU slightly higher than CNN-SFC, suggesting that OF-SFC is more accurate in terms of overlap between the predicted and actual labels. When considering the speed of processing, measured in FPS, CNN-SFC leads, which is notably faster than OF-SFC. This indicates that while OF-SFC may be more accurate, CNN-SFC offers a speed advantage.

Table II showcases the performance metrics of OF-SFC and CNN-SFC variants on four different subsets of the ZOD dataset. Notably, the CNN-SFC variant consistently achieves higher F1 Scores across all subsets, indicative of its superior precision and recall balance, particularly excelling in the $ZOD_2$ subset. This variant also maintains higher sensitivity, suggesting it is more adept at detecting true event occurrences. In contrast, the OF-SFC variant demonstrates remarkable specificity, particularly in $ZOD_1$ and $ZOD_2$, indicating its strength in correctly identifying non-events. Both variants present comparable mean IoU values across subsets, indicating similar accuracy in the spatial overlap of predicted and actual events. However, variability in processing speeds is observed for both OF-SFC and CNN-SFC, with CNN-SFC peaking in $ZOD_3$. This variability across the ZOD subsets can be attributed to the differing computational demands of each scenario.

> *Answering RQ-2:* When parameterizing our concept using synthetic data from SMIRK to subsequently evaluate it on the real-world dataset ZOD, we observe that the CNN-SFC variant generally achieves superior F1 Scores across all ZOD subsets, indicating a robust transfer of the learned model to real-world scenarios. CNN-SFC exhibits increased sensitivity, underscoring its potential in accurately detecting true events in a diverse array of real-world conditions. Meanwhile, the OF-SFC variant maintains commendable specificity, suggesting its applicability in scenarios where the accurate rejection of non-events is critical. Both variants demonstrate comparable spatial accuracy, yet their processing speeds are also comparable across the real-world dataset.

### C. Threats to Validity

*Internal Threats to Validity:* The OF-SFC and CNN-SFC variants were originally developed and fine-tuned on synthetic datasets, which could exhibit significant differences from real-world data in terms of variability and complexity. Without further calibration to bridge these differences, there is a risk that the variants may not maintain consistent performance when applied to diverse real-world scenarios.

*External Threats to Validity:* The ZOD dataset specifically captures European driving conditions, characterized by unique traffic patterns, driver behaviors, and environmental features. The results obtained from this dataset may not be fully representative of driving scenarios in other parts of the world. Hence, the OF-SFC and CNN-SFC variants may exhibit varied effectiveness when applied to new contexts.

*Generalizability:* The performance of the OF-SFC and CNN-SFC variants might be highly dependent on the specific characteristics of the SMIRK and ZOD datasets, which could limit the applicability of the results to other datasets with different attributes.

## VI. Conclusion and Future Work

In this paper, we focus on the systemtic detection of pedestrian crossings, using video data from a forward-facing camera mounted on a vehicle. We introduce and compare the performance of two different variants OF-SFC and CNN-SFC. SFCs allow for transforming multi-dimensional data obtained from the two variants into single-dimensional representations to enable event identification at scale. A large-scale synthetic dataset (SMIRK) and an extensive industrial real-world dataset (ZOD) are utilized to evaluate the two proposed variants. This assessment highlights their potential to enhance the performance and reliability of automotive perception systems across diverse traffic scenarios.

Future research may consider incorporating additional datasets from diverse geographical locations and varied environmental conditions to validate the robustness of the OF-SFC and CNN-SFC variants and confirm their effectiveness in a broader array of real-world scenarios. Additionally, further studies could investigate the integration of further sensor modalities to enhance the performance and reliability of these systems.

## Acknowledgments

This research is partially funded by the research project "SUNRISE", which has received funding from the European Union's Horizon 2020 Research & Innovation Actions under grant agreement No. 101069573, and by the Wallenberg AI Autonomous Systems and Software Program (WASP) funded by the Knut and Alice Wallenberg Foundation, and by the Swedish Research Council (VR) under grant agreement 2023-03810.

## Disclaimer

The views and opinions expressed are those of the authors and do not necessarily reflect the official policy or position of Volvo Cars. The proposed methods or the results generated by this work are only used in this study and not used in any engineering of production related projects.